\def\year{2019}\relax
\documentclass[letterpaper]{article} %DO NOT CHANGE THIS
\usepackage{aaai19}  %Required
\usepackage{times}  %Required
\usepackage{helvet}  %Required
\usepackage{courier}  %Required
\usepackage{url}  %Required
\usepackage{graphicx}  %Required
\usepackage{color}
\usepackage{multirow}
\usepackage{multicol}
\usepackage{booktabs}  
\usepackage{amssymb}
\usepackage{subfigure}

\definecolor{CQColor}{rgb}{0.0,0.0,1.0} % color for Aaron

\title{PVRNet: Point-View Relation Neural Network for 3D Shape Recognition}
\author{Haoxuan You\textsuperscript{1},\ Yifan Feng\textsuperscript{2},\ Xibin Zhao\textsuperscript{1},\ Changqing Zou\textsuperscript{3},\ Rongrong Ji\textsuperscript{2},\ Yue Gao\textsuperscript{1}\thanks{Corresponding author}\\ 
\textsuperscript{1}BNRist, KLISS, School of Software, Tsinghua University, China.\\
\textsuperscript{2}School of Information Science and Engineering, Xiamen University, China\\
\textsuperscript{3}UMIACS, University of Maryland, College Park, United States\\
\{haoxuanyou, evanfeng97, aaronzou1125\}@gmail.com, rrji@xmu.edu.cn, \{zxb, gaoyue\}@tsinghua.edu.cn, 
}
\begin{document}
\maketitle
\begin{abstract}
Three-dimensional (3D) shape recognition has drawn much research attention in the field of computer vision. The advances of deep learning encourage various deep models for 3D feature representation. For point cloud and multi-view data, two popular 3D data modalities, different models are proposed with remarkable performance. However the relation between point cloud and views has been rarely investigated. In this paper, we introduce Point-View Relation Network (PVRNet), an effective network designed to well fuse the view features and the point cloud feature with a proposed relation score module. More specifically, based on the relation score module, the point-single-view fusion feature is first extracted by fusing the point cloud feature and each single view feature with point-singe-view relation, then the point-multi-view fusion feature is extracted by fusing the point cloud feature and the features of different number of views with point-multi-view relation. Finally, the point-single-view fusion feature and point-multi-view fusion feature are further combined together to achieve a unified representation for a 3D shape. Our proposed PVRNet has been evaluated on ModelNet40 dataset for 3D shape classification and retrieval. Experimental results indicate our model can achieve significant performance improvement compared with the state-of-the-art models.    
\end{abstract}

\section{Introduction}

With the proliferation of image and 3D vision data, 3D shape recognition has grown into an important problem in the field of computer vision with a wide range of applications from AR/VR to self-driving. Several modalities are used to represent 3D shape and the recent development of deep learning enables various deep models to tackle different modalities. For volumetric data, 3D convolutional neural networks (CNN) \cite{maturana2015voxnet}\cite{wu20153d}\cite{qi2016volumetric} are generally used, but the sparse structure of volumetric data and the high computation cost constrain the performance. For view based and point cloud based models, the input data respectively are multiple views from cameras and point clouds from sensors and LiDARs, which have rich information and can be easily acquired. More and more research interests are concentrated on dealing with multi-view data and point cloud data.

\begin{figure}[!tbp]
  \centering
  \includegraphics[width=3.2in]{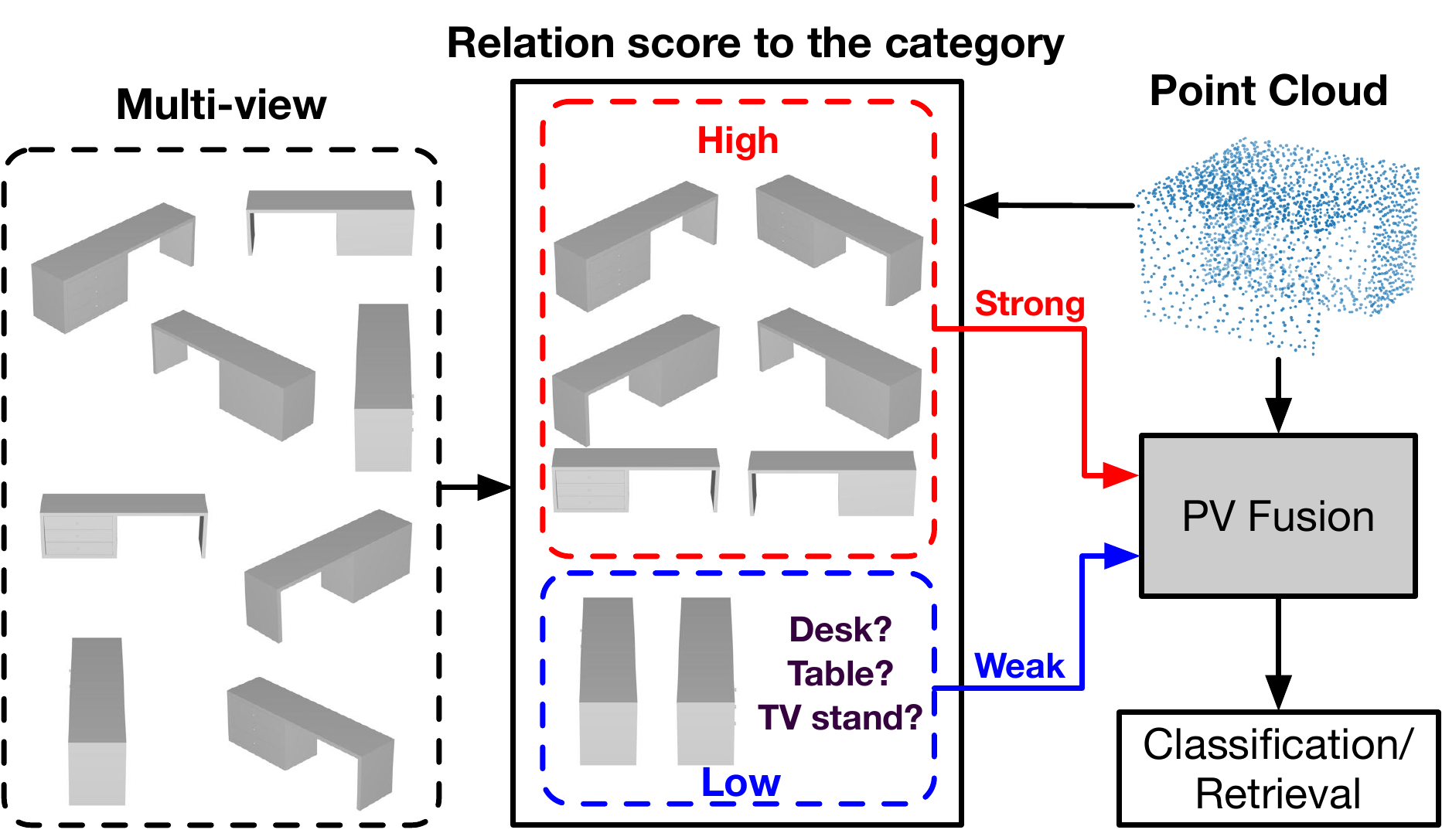}
  \caption{Illustration of the motivation of PVRNet.}
\label{fig:overview}
\end{figure}

\begin{figure*}[!htbp]
  \centering
  \includegraphics[width=6.5in]{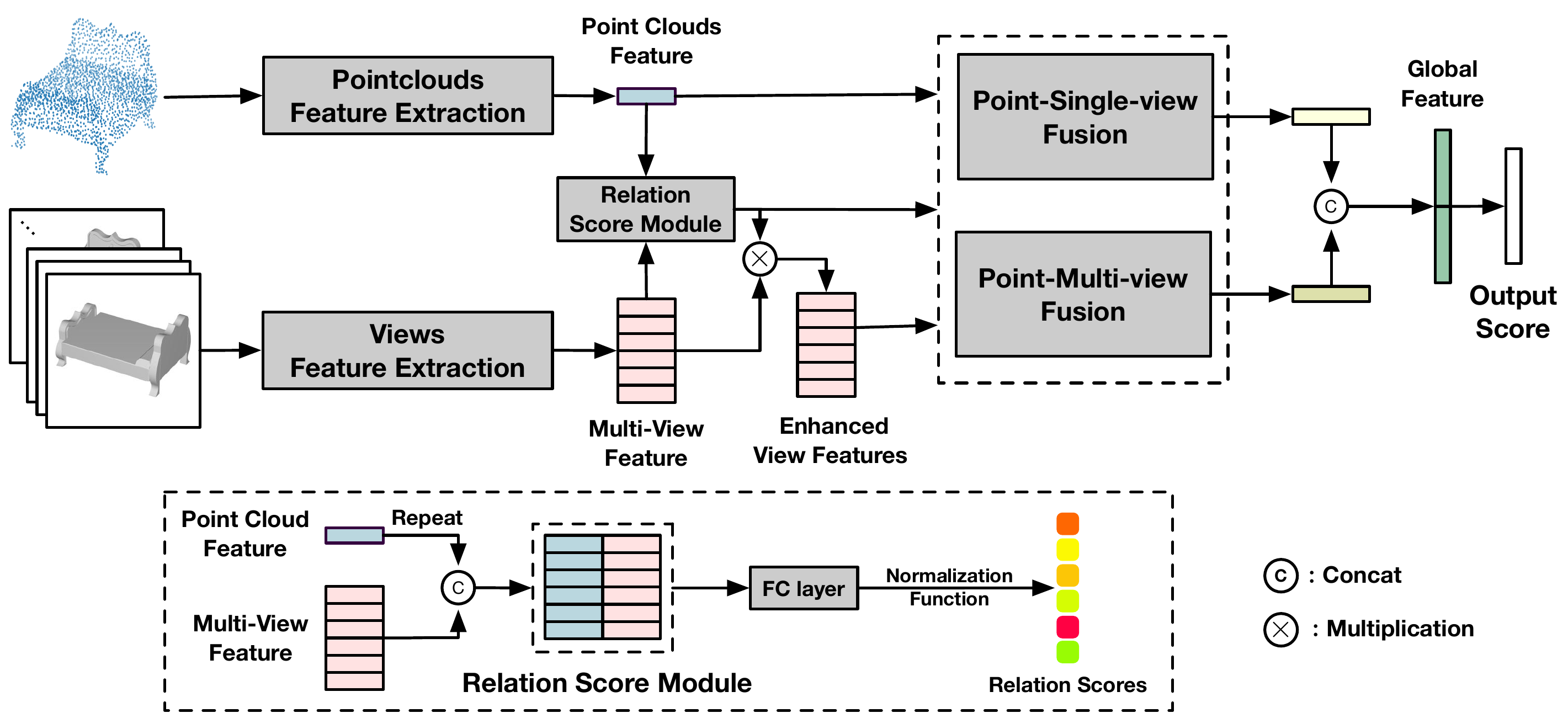}
  \caption{Architecture of the proposed PVRNet.}
\label{fig:pipline}
\end{figure*}

For view based deep models, multiple views captured by cameras with different angles are employed to represent 3D shape as the input. These views can be efficiently processed with 2D CNN \cite{su2015multi} \cite{feng2018gvcnn}. However, limited by camera angles, each view only presents partial local structures of an entire 3D shape, and thus some view features may not be discriminative for shape recognition. For point cloud based deep models, the input data are irregular point clouds obtained from 3D sensors, which well reserves the original spatial structure by 3D coordinates. While conventional CNN is not suitable for disordered data, making extracted feature ineffective. 

Considering the pros and cons of the view based models and point cloud based models, the fusion of point cloud and multi-view features may obtain a better representation for 3D shape. Existing work \cite{you2018pvnet} fuses the global view feature after view-pooling with the point cloud feature for 3D shape recognition. However, multi-view features are equally treated when combined with the point cloud feature. We argue that this kind of fusion cannot fully exploit the relation between multi-view features and the point cloud feature. Our point is based on the fact that each view presenting various amount of local structures certainly brings different effect in the feature fusion process with the point cloud feature. Generally, the views with discriminative descriptors are more beneficial for the fused 3D shape representation than the view descriptors with poor discrimination. Therefore, it is essential to mine the relation between the point cloud and different views to achieve a more powerful shape descriptor.

In this work, we propose a point-view relation neural network called PVRNet, which serves as the first exploration on the relationship between point cloud an multi-view data. At First, a relation score module which computes the relation scores between point cloud and each view is introduced. And the point-single-view fusion is performed with the point-single-view relation. Second, based on the relation scores, point-view sets (PVSets) which containing the point cloud and the different number of views are fused together to form the point-multi-view fusion feature, which exploits the point-multi-view relation. In this process, only views having higher relation scores are included in PVSets. In such way, more informative view features are preserved and combined with the point cloud feature. Third, the point-single-view fusion feature and the point-multi-view fusion feature are further fused together to obtain a unified 3D shape representation. To demonstrate the effectiveness of the proposed PVRNet, we have performed experiments on ModelNet40~\cite{wu20153d}, the most popular 3D shape dataset, for the tasks of shape classification and retrieval. The experimental results show our framework can outperform not only existing point cloud based or view based methods but also multimodal fusion methods. The evaluation results confirm that the point-view relation based feature fusion scheme achieves a powerful 3D shape representation.

The main contributions of this paper are two-fold:

(1) We introduce the first point-view relation (relevance) based deep neural network to well fuse the multi-view data and point cloud data. Different from existing work treating views equally for the feature fusion, our network fuses the two-modality features by exploiting the correlation between point cloud and each individual view by the relevance (relation score). 

(2) We design a new relation fusion scheme for the point cloud feature and the view features. In our fusion scheme, two kinds of relations including point-multi-view relation and point-singe-view relation are exploited to well fuse the point cloud feature and the view features together to form an integrated 3D shape representation.

The rest of our paper is organized as follows. We first present the related work. Then we detail the proposed PVRNet. After that, we show the experimental results as well as provide some insight-full analysis. In the last section, we conclude the paper.

\section{Related Work}
In this section, we briefly review some existing related methods, which can be divided into three directions: view based methods, point clouds based methods and multimodal methods. 
\subsubsection{View Based Methods}
For view based models, the input data is a group of views taken from cameras with different angles. Hand craft methods are explored at first. Lighting Field Descriptor is a typical view based 3D shape descriptor, whose input is a group of ten views and is obtained from the vertices of a dodecahedron over a
hemisphere. With the proliferation of deep learning, many deep view based models employing deep neural networks are proposed. As a classic model, a multi-view convolutional neural network (MVCNN) is introduced in \cite{su2015multi}, where views are fed into a weight-shared CNN and followed by a view-pooling operation to combine them. In GVCNN \cite{feng2018gvcnn}, the relationships between different views are exploited by a group based module. In 3D shape retrieval task, various models are introduced. In \cite{gao2012camera}, the 3D objects retrieval is performed by the probabilistic matching. In \cite{guo2016multi}, the complex intra-class and inter-class variations are settled by a deep embedding network supervised by both classification loss and triplet loss. In \cite{xie2017deepshape}, a deep network with auto-encoder structure is proposed for 3D feature extraction. Triplet-center loss is introduced in \cite{he2018triplet} as a well-designed loss function especially for retrieval. In \cite{dai2018siamese}, a bidirectional long short-term memory (BiLSTM) module is employed to aggregate information across different views.
\subsubsection{Point Clouds Based Methods}
For point cloud based models, the input data is a set of points sampled from the surface of the 3D shape. Although point cloud preserves more completed structure information, the irregular and unstructured data prevent the usage of conventional 2D CNN. As a pioneer, PointNet \cite{qi2017pointnet} first introduces a deep neural network to process disordered point cloud data. To be invariant to the permutation, spatial transform block and a symmetric function-max pooling are applied. Following PointNet++ is proposed in \cite{qi2017pointnet++} to employ PointNet module in local point sets and then the local features are aggregated in a hierarchical way. In Kd-Netowrk \cite{klokov2017escape}, features are extracted and gathered by the subdivision of points on Kd-trees. For better exploiting local structure information, DGCNN in \cite{wang2018dynamic} proposes EdgeConv operation to get edge feature between points and their neighbors. In PointCNN \cite{li2018pointcnn}, the $\chi\mbox{-Conv}$ operation is proposed to extract local patch features and a hierarchical structure containing several $\chi\mbox{-Conv}$ layers is applied like traditional CNN models.
\subsubsection{Multimodal Methods}
Compared with 2D image, 3D shape usually has more complex structure information, which makes a single modality difficult to describe the 3D shape completely. So the multimodal fusion is necessary and beneficial if in an effective way\cite{gao2018question}. In \cite{hegde2016fusionnet}, FusionNet jointly employs volumetric data and view data together to learn a unified feature representation. When it comes to the fusion of view data and point cloud data, PVNet \cite{you2018pvnet} uses the global view feature to guide the local feature extraction of point cloud. In the task of 3D object detection, MV3D network \cite{chen2017multi} fuses the point cloud data from LiDAR and view data from camera together to perform detection task, which however projects the point cloud data to bird's eye view map and perform fusion in image level. Our work differs greatly from them by mining the relationships between point cloud data and multi-view data and proposing the point-single-view fusion and point-multi-view fusion based on the relation of two modalities.

\section{Point-View Relation Network}

\begin{figure}[!htbp]
  \centering
  \includegraphics[width=3.4in]{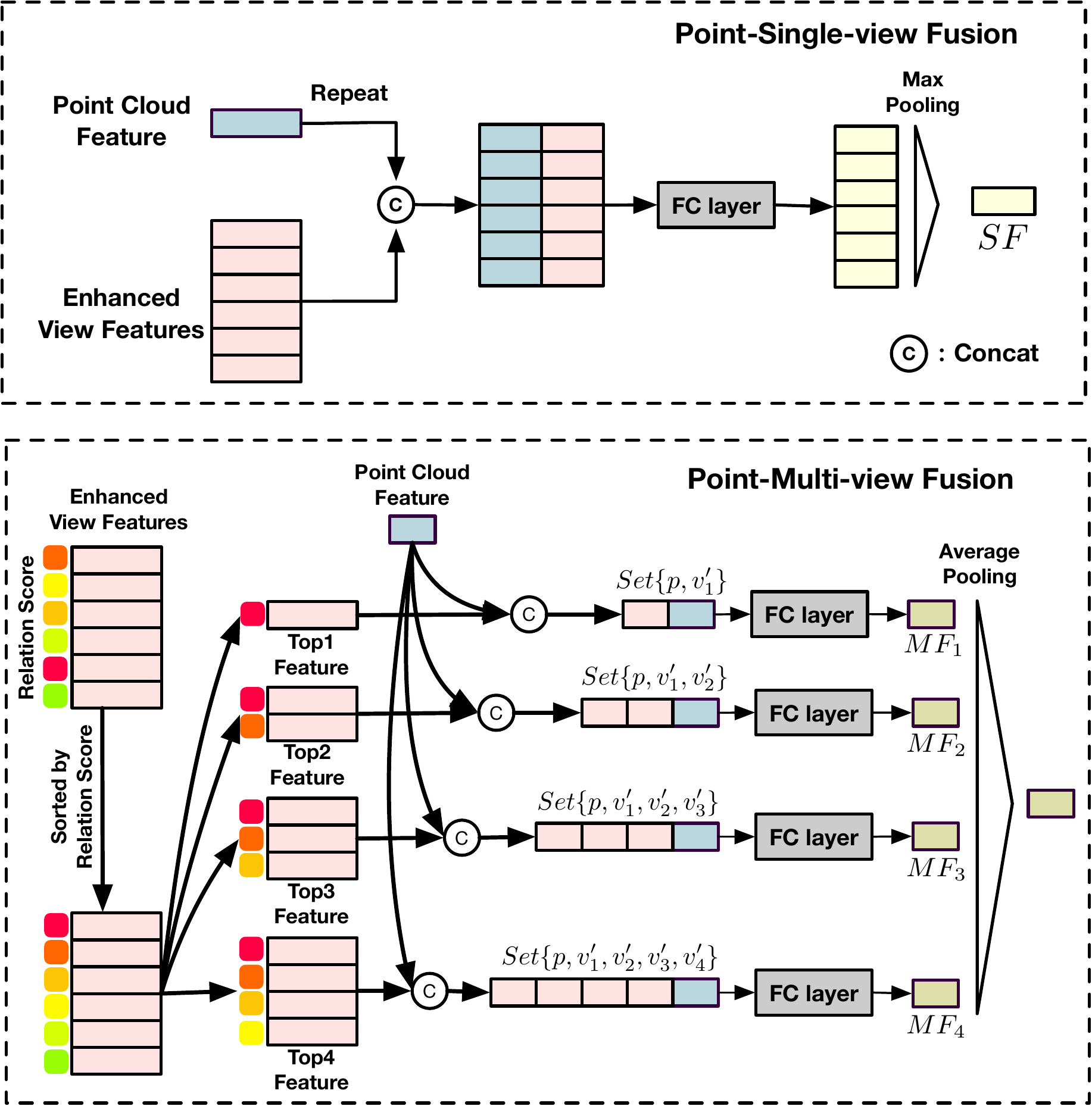}
  \caption{Point-multi-view fusion and point-single-view fusion modules. The enhanced view features are sorted by the relation scores. Different colors denote different significance values of relation scores. From the red to green, the relation scores decrease from 1 to 0.
  }
\label{fig:sub_block}
\end{figure}

In this section, the details of our proposed PVRNet are introduced. The input of PVRNet are two modality data of a 3D shape: multi-view and point cloud data. The data acquisition setting follows \cite{you2018pvnet} to get 12 views and 1,024 points for each shape. Point cloud and multi-view data are first individually fed to the corresponding feature extractor to get the global feature of the whole point cloud and the view features for each view. Then, a relation score module is proposed to learn the relevance of the point cloud and each view. Based on the relation scores, both point-single-view fusion and point-multi-view fusion are further exploited to build a unified feature for 3D shape. The unified feature is then used for the applications of shape classification and retrieval. The architecture of the whole framework is shown in Fig. \ref{fig:pipline}

\subsection{Feature Extraction}
\subsubsection{Point cloud} 
Let a point cloud of $F$-dimension with $n$ points be $X = \{x_1, \ldots, x_n\}\subseteq \mathbb{R}^F$. In the general setting of $F=3$, points are presented with 3D coordinates. We employ DGCNN \cite{wang2018dynamic} to extract the global feature from point cloud. DGCNN includes a 3D spatial transform network, several EdgeConv layers, and a max-pooling to aggregate features. Note that our framework is compatible with various models for point cloud feature extraction.

\subsubsection{Multi-view} 
Each 3D shape is represented by a group of rendered views captured by a predefined camera array. Following MVCNN \cite{su2015multi}, we feed these views to a group of CNNs that share the same parameters. The view-pooling is discarded so that the features of each view are preserved.

\subsection{Relation Score Module}
For multimodal representation tasks, the core task is how to effectively fuse features from multiple streams. In multi-view branch, each view owns a part of local structures and view features have various levels of discriminability for shape representation. When fused with point cloud features, all view features cannot be integrated effectively by a simple equal pooling scheme. Hence, it is essential to mine the relation between point cloud and each view so as to find out an effective
solution for feature fusion. 

Inspired by the relational reasoning network for the VQA task \cite{santoro2017simple}, we define the relation score of point cloud and $i^{th}$ view as:
\begin{equation}
  RS_i(P,V) = \xi(g_\theta(p, v_i)),
\end{equation}
where $p$ is the point cloud feature and $V = \{v_1,\ldots,v_n\}$ denotes $n$ extracted view features from a 3D shape. The function $g_\theta$  reasons the relations between the point cloud feature and per-view feature and learns an effective fusion. In our network, $g_\theta$ uses the simple multilayer perceptrons (MLP) and $\xi$ is the normalization function (sigmoid function is used in our implementation). For each view, the output is a relation score ranging from 0 to 1, which represents the significance of the correlation between different views and the point cloud. 

The relation scores have two purposes: 1) enhance the view features in a residual way and 2) decide which view feature will be included in point-multi-view fusion. For feature enhancement, the view features that have a stronger correlation with the point cloud feature will be assigned a larger significance. We use the relation score $RS(P,V)$ to enhance the view features by a residual connection:
\begin{equation}
  v_i' = v_i*(1+RS_i(P,V))
\end{equation}
where $v_i*RS_i(P,V)$ is to refine the feature by relation score and is then added to origin view feature $v_i$ to generate the enhanced feature $v_i'$. After that, the enhanced view feature, point cloud feature, and relation scores are sent into the point-single-view fusion module and point-multi-view fusion module to get a integrated feature. The point-single-view fusion and point-multi-view fusion are illustrated in Fig. \ref{fig:sub_block}

% \subsection{Multi-Set Relation Fusion}
% Each view has its own local feature, so we further capture the relationships between the point cloud and different number of views. Consider a point-view set $S_n=\{p,v'_1,\ldots,v'_n\}$, where $n$ denotes the number of views in this set and $p$ denotes the corresponding point cloud feature. According to the number of views, point-single-view relation or point-multi-view relation are explored and the features of multiple PVSets are fused together to achieve an integrated representation. The point-single-view relation and point-multi-view relation are illustrated in Fig. \ref{fig:sub_block}
\subsection{Point-Single-View Fusion}
Each view has its own local feature, so it is reasonable to employ the fusion between the point cloud and different views. Consider a point-view set (PVSet) $S_n=\{p,v'_1,\ldots,v'_n\}$, where $n$ denotes the number of views in this set and $p$ denotes the corresponding point cloud feature. According to the number of views in PVSets, point-single-view fusion and point-multi-view fusion are performed. Point-single-view fusion module takes $n$ PVSets containing point cloud and each view as input and aggregates the pairwise relations.

When the PVSet only includes single view $i$ , $S_{1i} = \{p, v'_i\}$. The pairwise set fusion function can be defined as:
\begin{equation}
  SF_{i} = h_\phi(p, v'_{i}),
\end{equation}
where the subscript $i$ denotes that the fusion is about a point-single-view  set with the $i^{th}$ view. We use multilayer perceptrons (MLP) as function $h_\phi$. For point-single-view set, only one set can not provide enough information. We therefore aggregate $n$ set relations covering all views with a simple max-pooling operation. The final output of this block can be described as below:
\begin{equation}
  SFusion = SF = Maxpooling\{SF_{1},\ldots,SF_{n}\},
\end{equation}
\subsection{Point-Multi-View Fusion}
We further explore the fusion between point cloud and multiple views. Under this circumstance, point cloud is combined with several views to form PVSets. Relation scores here decide which view features are included in the sets. The view features are sorted according to the values of the relation scores. Then the view features with higher relation scores are selected to combine point cloud feature to form PVSets. A set with $k$ views can be denoted as $S_k = \{p, v'_1, \ldots, v'_k\}$, where the top-$k$ views that have the highest relation scores are included. The fusion function can be extended from the point-single-view fusion above:
\begin{equation}
  MF_k = h_\phi'(p, v_1', \ldots,v_k').
\end{equation}
With total $K$ PVSets, different point-multi-view combinations are accumulated together:
\begin{equation}
  MFusion = \frac{1}{K}\sum_{k=1}^{K}MF_k.
\end{equation}

After that, the fusion features of two levels are aggregated together by a concatenation operation to generate the final point-view relation feature:
\begin{equation}
  Fusion = Concat(SFusion, MFusion),
\end{equation}
where $SFusion$ denotes point-single-view fusion feature and $MFusion$ denotes point-multi-view fusion feature. Then the final feature is followed by MLPs and softmax function to produce the classification result. 

% \begin{figure}[!tbp]
%   \centering
%   \includegraphics[width=3.2in]{LaTeX/sub_block.pdf}
%   \caption{The point-single-view relation module and point-multi-view relation module.
%   }
% \label{fig:sub_block}
% \end{figure}

% \subsection{Implementation}
% \subsubsection{Network Architecture}
% In our PVRNet framework, point cloud feature and view feature extraction can be performed by any view based and point cloud based models. In our implementation, the 12 view features are extracted by 12 AlexNets that share the same paramters like \cite{su2015multi}. But no view-pooling is performed and the feature of each view is preserved. For point cloud feature extraction, DGCNN\cite{wang2018dynamic} is employed.
\subsection{Implementation and Training Strategy}
In our PVRNet framework, point cloud feature and view feature extraction can be produced by any view based and point cloud based models. In our implementation, the 12 view features are extracted by 12 AlexNets that share the same parameters like \cite{su2015multi}. No view-pooling is performed to preserve the feature of each view. PVRNet is trained in an end-to-end fashion. The view feature extraction model and point cloud feature extraction model is initialized by the pre-trained MVCNN model and DGCNN model. And an alternative optimization strategy is adopted to update our framework. In first 10 epochs, the feature extraction model is fixed and we only fine-tune the other part (including relation score module, point-single-view fusion module and point-multi-view fusion module). After the first 10 epochs, the all parameters are updated together to get better performance. The intuition of this strategy is that the feature extraction part is well-initialized but the relation module and fusion modules are relatively weak in the beginning, and this strategy is more stable for the fusion. 
% \subsubsection{Classification and Retrieval Setting}

% The big table
\begin{table*}[!htbp]
\begin{center}
\begin{tabular}{lccc}
\toprule
\multirow{2}{*}{Method}& Data Representation. & Classification  & Retrieval  \\
\cline{2-2}
   &Number of Views & (Overall Accuracy) & (mAP) \\
\midrule
(1) 3D ShapeNets \cite{wu20153d} & Volumetric & 77.3$\%$ & 49.2$\%$ \\
(2) VoxNet \cite{maturana2015voxnet} & Volumetric & 83.0$\%$ & - \\
(3) VRN \cite{brock2016generative} & Volumetric & 91.3$\%$ & - \\
(4) MVCNN-MultiRes \cite{qi2016volumetric} & Volumetric & 91.4$\%$ & - \\
\midrule
(5) LFD \cite{chen2003visual} & 10 Views & 75.5$\%$ & 40.9$\%$ \\
(6) MVCNN (AlexNet) \cite{su2015multi} & 12 Views & 89.9$\%$ & 80.2$\%$ \\
(7) MVCNN (GoogLeNet)& 12 Views& 92.2$\%$ & 83.0$\%$ \\

(8) GVCNN (GoogLeNet) \cite{feng2018gvcnn}& 8 Views& 93.1$\%$ & 84.5$\%$ \\
(9) GVCNN (GoogLeNet) & 12 Views& 92.6$\%$ & 85.7$\%$ \\
% (10)RotationNet \cite{kanezaki2016rotationnet}& 12 Views& 90.65 & -$\%$ \\
(10) MVCNN with TCL \cite{he2018triplet}& 12 Views& - & 88.0$\%$ \\

\midrule

(11) PointNet \cite{qi2017pointnet} & Point Cloud & 89.2$\%$ & - \\
(12) PointNet++ \cite{qi2017pointnet++} & Point Cloud & 90.7$\%$ & - \\
(13) KD-Network \cite{klokov2017escape} & Point Cloud & 91.8$\%$ & - \\
(14) SO-Net \cite{li2018so} & Point Cloud & 90.9$\%$ & - \\
(15) DGCNN \cite{wang2018dynamic} & Point Cloud & 92.2$\%$ & 81.6\% \\
\midrule

(16) FusionNet \cite{hegde2016fusionnet} & Volumetric and 20/60 Views& 90.8$\%$ & - \\
(17) PVNet \cite{you2018pvnet} & Point Cloud and 12 Views & 93.2$\%$ & 89.5$\%$ \\
\midrule

(18) \textbf{PVRNet} (AlexNet) & Point Cloud and 12 Views & \textbf{93.6}$\%$ & \textbf{90.5}$\%$ \\
\bottomrule
\end{tabular}
\end{center}
\caption{Classification and retrieval results on the ModelNet40 dataset. In experiments, our proposed framework PVRNet is compared with state-of-the-art models that use different representations of 3D shapes. MVCNN (GoogLeNet) means that GoogLeNet is employed as base architecture for weight-shared CNN in MVCNN. PVRNet (AlexNet) indicates using AlexNet as base structure in our view feature extraction and our PVNet framework can get superior performance over other models.}
\label{tab:experiments}
\end{table*}

\section{Experiments}
In this section, the experiment results of PVRNet on the tasks of classification and retrieval are presented. The analysis of results and comparison with the state-of-the-art methods including both point cloud methods, multi-view methods, and multimodal methods are also provided. In addition, 
the ablation study of the proposed point-view relation fusion module is also investigated. At last, we explore the effect of different number of views and points toward the classification performance.

\begin{figure}[!htbp]
  \centering
  \includegraphics[width=3.2in]{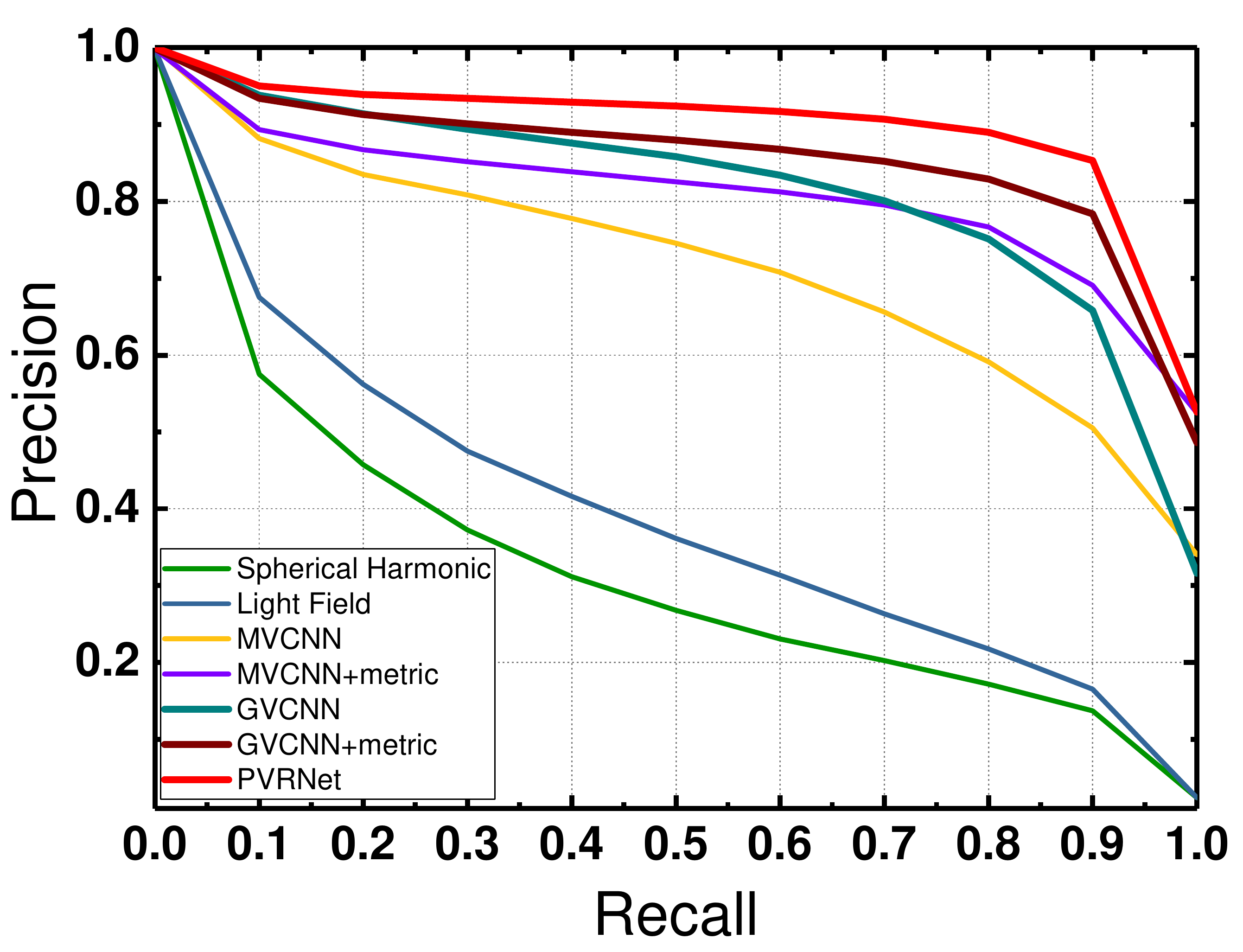}
  \caption{The precision-recall curves of our PVRNet and other compared methods in the task of shape retrieval. In MVCNN and GVCNN models, GoogLeNet is employed as the base network. The metric denotes the low-rank Mahalanobis metric learning is used. Our PVRNet, with no specific metric and retrieval loss employed, still get superior performance over others.}
\label{fig:pr}
\end{figure}

\subsection{3D Shape Recognition and Retrieval}

To demonstrate the performance of the proposed PVRNet on 3D shape recognition and retrieval, comprehensive experiments are conducted on the dataset of Princeton ModelNet \cite{wu20153d}. ModelNet contains 127,912 3D CAD models from 662 categories. ModelNet40, a more commonly used subset containing 12,311 3D CAD models from 40 popular categories, is applied in our experiments. Our training and test split setting follows \cite{wu20153d}. Individually, point cloud data are sampled from the surface of each CAD model as \cite{qi2017pointnet} and view data are captured by camera as the setting in \cite{su2015multi}.

In experiments, various methods based on different modalities are compared with PVRNet. Volumetric based methods such as 3D ShapeNets\cite{wu20153d}, VoxelNet\cite{maturana2015voxnet}, VRN\cite{brock2016generative}, and MVCNN-MultiRes\cite{qi2016volumetric} are compared. Multi-view based methods consist of hand-craft descriptor LFD\cite{chen2003visual}, deep learning models MVCNN\cite{su2015multi}, and GVCNN\cite{feng2018gvcnn}. PointNet\cite{qi2017pointnet}, PoineNet++\cite{qi2017pointnet++}, Kd-Network\cite{klokov2017escape}, DGCNN\cite{wang2018dynamic} and SO-Net\cite{li2018so} are included in the point cloud based methods. For multimodal methods, FusionNet\cite{hegde2016fusionnet} and PVNet\cite{you2018pvnet} are compared. 

% For fair comparison, VRN \cite{brock2016generative} is presented with the result of single CNN and the result of RotationNet \cite{kanezaki2016rotationne} is reported with views captured by the default camera arrays identical to other methods \cq{[not clear here]}.  

The experimental comparison is shown in Tab.~\ref{tab:experiments}. PVRNet achieves the best performance with the classification accuracy of 93.6\% and retrieval mAP of 90.5\%. Compared with view based models, our proposed PVRNet with AlexNet has an improvement of 1.4\% and 7.5\% over the classic MVCNN (use GoogLeNet as backbone) on the classification and retrieval tasks, respectively. PVRNet also outperforms the state-of-the-art GVCNN with the improvement of 0.5\% and 4.8\% on the classification and retrieval tasks, respectively. Compared with state-of-the-art point cloud based models, PVRNet outperforms DGCNN by 1.4\% and 8.9\% in terms of classification and retrieval accuracy, respectively. For multimodal methods, the performance of PVRNet also outperforms others.

For retrieval task, we take the 256-dimensional feature before the last fully-connected layer as the shape representation. For MVCNN and GVCNN, a low-rank Mahalanobis metric learning \cite{su2015multi} is applied to boost the retrieval performance. As in \cite{he2018triplet}, triplet-center loss is used as the loss function. Trained with general softmax loss and without low-rank Mahalanobis metric, PVRNet can achieve the state-of-the-art retrieval performance with 90.5\% mAP, which indicates the capability of the fused feature for 3D shape representation. Fig. \ref{fig:pr} presents the precision-recall curve of our model. It is shown that our PVRNet, without specific metric learning and retrieval loss, can still significantly outperforms the point cloud based model and view based methods that have triplet-center loss and metric learning.
\subsection{Ablation Analysis}

\begin{table}[!tpb]
  \begin{center}
  \begin{tabular}{ccc}
    \toprule
    \multirow{2}{*}{Models} & Mean Class&Overall\\
     & Acc. & Acc. \\
    \midrule
    Point Cloud Model& 90.2\% & 92.2\%\\
    Multi-View Model & 87.6\% & 89.9\%\\
    Late Fusion & 90.81\% & 92.62\%\\
    SFusion & 90.93\% & 92.84\%\\
    S+M Fusion(1+2view) & 91.35\% & 92.99\%\\
    S+M Fusion(1+2+3view)  & 91.43\% & 93.47\%\\
    S+M Fusion(1+2+3+4view) & \textbf{91.64}\% & \textbf{93.61}\%\\
  \bottomrule
\end{tabular}
\end{center}
  \caption{Ablation experiments of our framework on the classification task.}
  \label{tab:ablation}
\end{table}

\begin{table}[!htbp]
  
  \begin{center}
  \begin{tabular}{cccc}
    \toprule
    \multirow{2}*{Models}&Number of&Mean Class&Overall\\  
    ~& Views & Acc & Acc\\
    \midrule
    \multirow{4}*{MVCNN}&4& 82.5\% & 84.6\%\\
    ~&8& 87.2\% & 89.0\%\\
    ~&10& 87.4\% & 89.3\%\\
    ~&12& 87.6\% & 89.9\%\\
    \midrule
    \multirow{4}*{GVCNN}&4& 88.1\% & 90.3\%\\
    ~&8& 90.2\% & 92.1\%\\
    ~&10& 90.4\% & 92.4\%\\
    ~&12& 90.7\% & 92.6\%\\
    \midrule
    \multirow{4}*{PVRNet}&4& 89.5\% & 91.5\%\\
    ~&8& 91.4\% & 93.2\%\\
    ~&10& 91.5\% & 93.5\%\\
    ~&12& 91.6\% & 93.6\%\\
  \bottomrule
\end{tabular}
\end{center}
\caption{The comparison of different number of input views. 1,024 points and 12 views are feed to train models and we respectively employ 4, 8, 10, 12 views as our test data that are captured by cameras with 90$^\circ$, 45$^\circ$, 36$^\circ$, 30$^\circ$ interval. Top: The result of MVCNN; Middle: The result of GVCNN; Bottom: The result of our PVRNet.}
  \label{tab:view_miss}
\end{table}

In this section, we go deep into PVRNet to investigate the effectiveness of our point-view relation fusion method. There are two popular fusion methods, early fusion and late fusion, for multimodal fusion task. For the problem of the fusion of point cloud and multi-view in this paper, early fusion is not available because of the different dimension number of two representations as well as the disordered property of point cloud. However, as for late fusion, the global feature from point cloud model and global view feature after the step of view-pooling can be directly concatenated together for shape representation. But various views have different relationships with point cloud. Some views are more informative for combining with point cloud feature. So in our PVRNet, the relations between point cloud and views are described by relation scores, which then guide point-view fusion to build a unified representation for 3D shape. 

To well study the effectiveness of PVRNet, we evaluate the performance of different combinations of our model's components. And the influence of different number of PVsets is investigated. The detailed experimental results are shown in Tab. \ref{tab:ablation}. In the table, "Point Cloud Model" denotes the point cloud feature extraction model used in PVRNet. "Multi-View Model" denotes the multi-view feature extraction model we use where MVCNN with AlexNet \cite{krizhevsky2012imagenet} is employed. "Late Fusion" denotes the result of direct late fusion of two global features. "SFusion" denotes only Point-single-view fusion block is employed. "S+M Fusion(1+...+ $n$ view)" denotes employing both point-single-view fusion and point-multi-view fusion with $n$ sets, where each set includes point cloud and $n$ views. PVRNet outperforms not only the baseline multi-view and point cloud methods by a large margin but also the late fusion method by about 1\% in overall accuracy and 0.8\% in mean class accuracy. And it can be seen that our point-single-view fusion and point-multi-view fusion are both beneficial to the performance. With more meaningful views included in point-multi-view fusion block, the performance of our network progressively improves from 92.84\% to 93.61\%, which validates the effectiveness of our point-multi-view fusion.
% The relations between point cloud and different numbers of views can be effectively captured and benefit the shape recognition. 

\begin{figure*}[!htbp]
  \centering
  \includegraphics[width=7in]{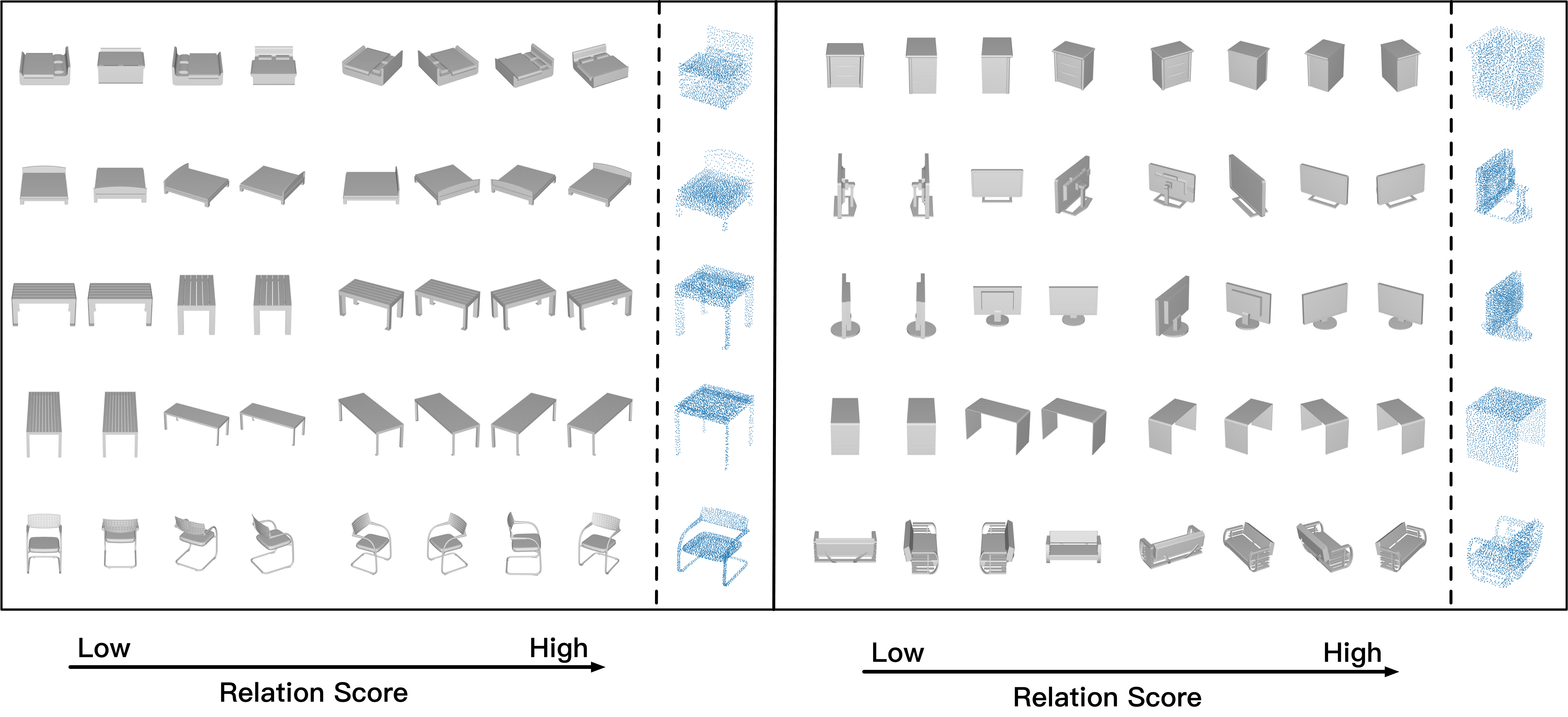}
  \caption{Visualization of relation score module. For each 3D shape, we list the views with relation scores from low to high value and corresponding point cloud samples. }
\label{fig:vis}
\end{figure*}

\begin{figure}[!htbp]
  
    \begin{center}  
    \subfigure[Overall accuracy]{
    \includegraphics[width=1.56in]{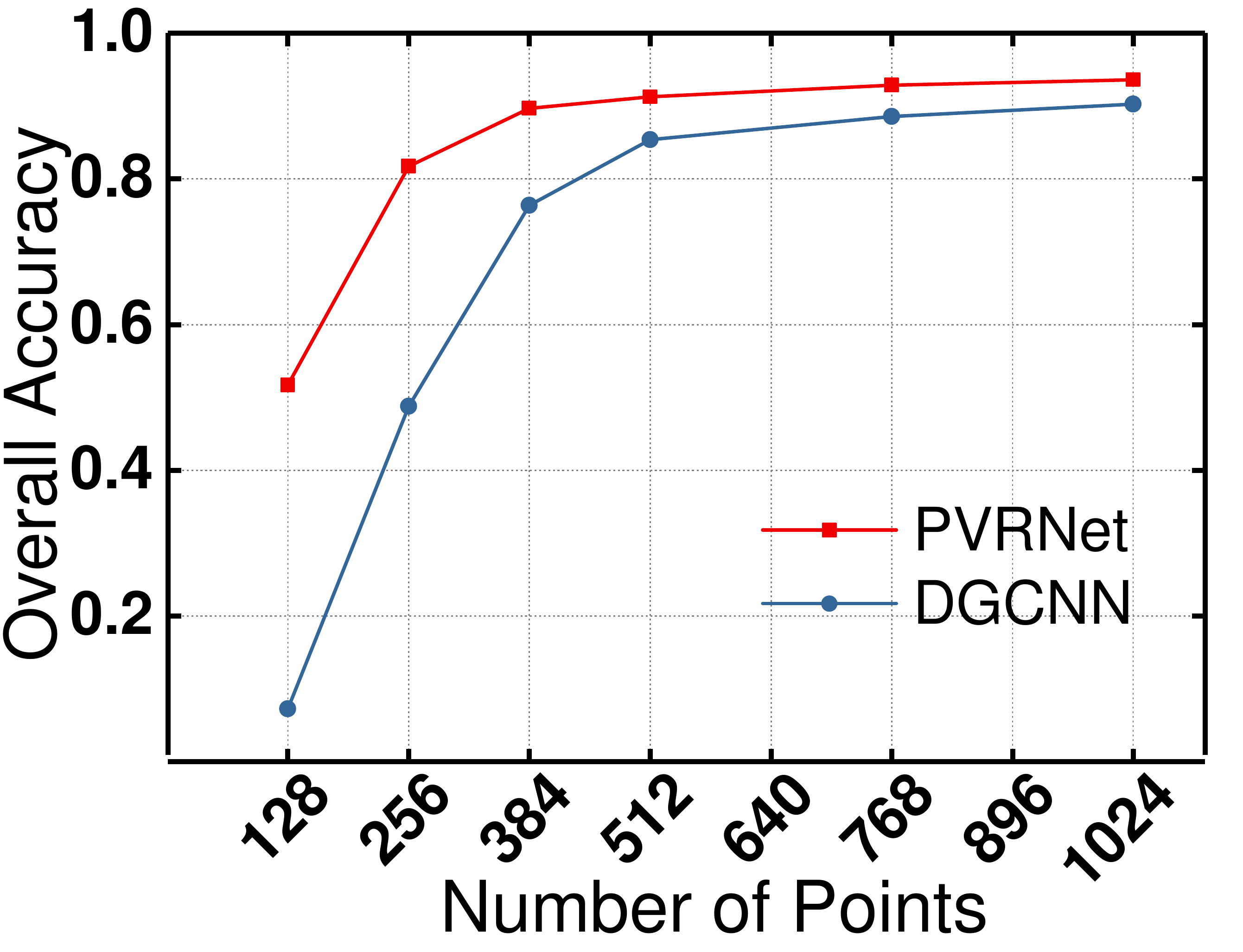}}
    \subfigure[Mean class accuracy]{
    \includegraphics[width=1.56in]{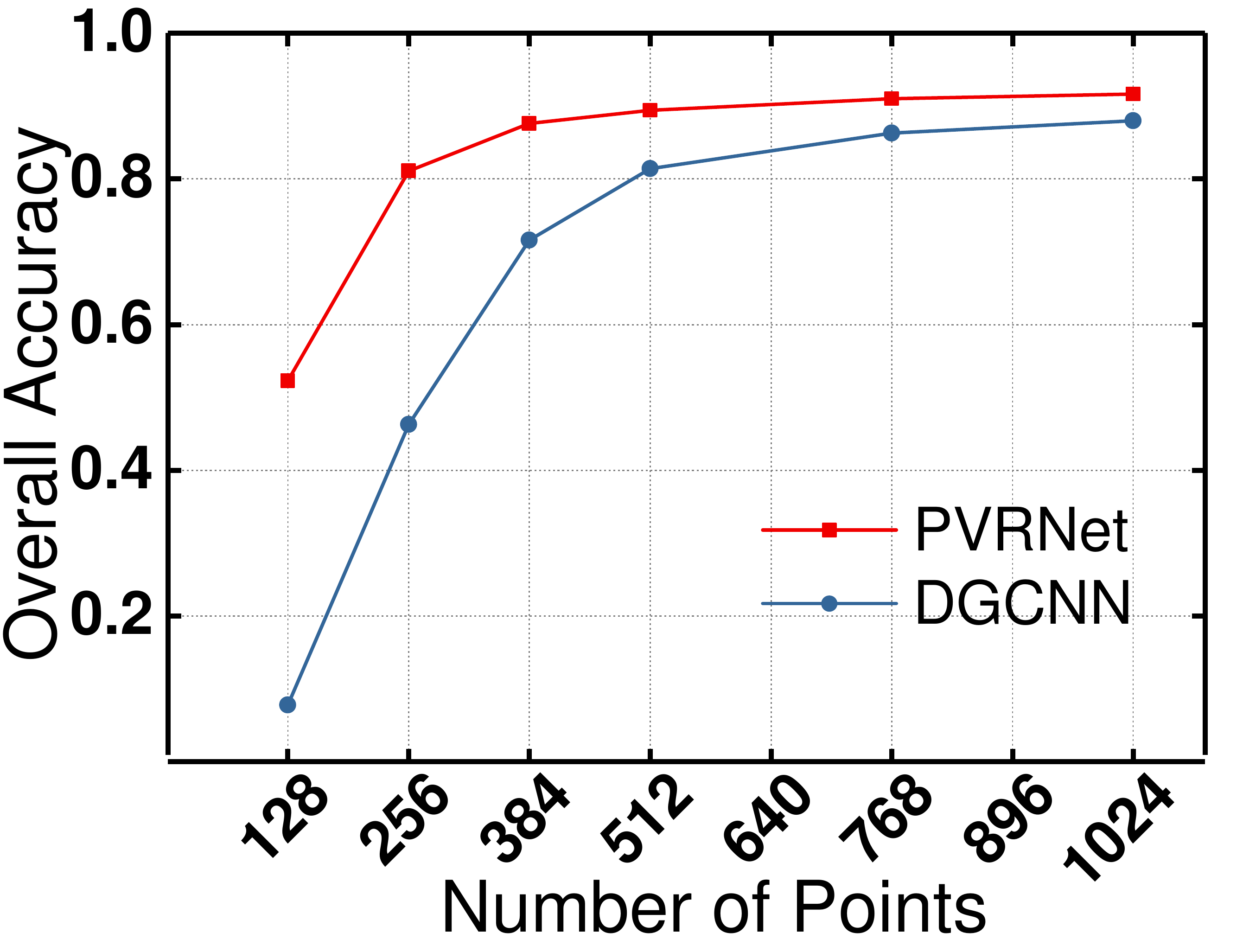}
    }
    \end{center}
    
  \caption{The experimental result of different number of input points. Left: Overall accuracy of PVRNet and DGCNN. Right: Mean class accuracy of PVRNet and DGCNN.}
\label{fig:rubust_of_point_num}
\end{figure}

To better explain the effect of our PVRNet, we further visualize the views sorted by relation scores in Fig.\ref{fig:vis}. It can be found that the views having higher relation scores are usually more informative, which are beneficial to be combined with point cloud. So our relation score module can effectively capture the correlations between point cloud and views and help to build and powerful feature for shape representation.
%So our PVRNet can efficiently capture the correlations between point cloud and different numbers of views and obtain an effective feature for shape representation.

\subsection{Influence of Missing Data}

In this section, we evaluate the robustness of PVRNet toward missing data. In real world, it often happens that we can only get part of data as input. At that time, we also expect the network to give satisfactory performance under this situation. It can demonstrate the generalization ability and robustness of our method.

At first, we investigate the influence of missing views. PVRNet is trained with 12 views and 1024 points. In the testing, we keep the number of points fixed as 1,024 but vary the number of views. 4, 8 10, 12 views are successively selected as input. The comparison of PVRNet and MVCNN and GVCNN  is shown in Tab. \ref{tab:view_miss}. Note that MVCNN and GVCNN are also trained with 12 views. It indicates that with the reduced views, PVRNet can still keep a high performance of more than 91\% accuracy. And with 4 views, compared with the performance drop of 5.3\% of MVCNN and 2.3\%, PVRNet only drops 2.1\%. 

Then we explore the influence of missing points. In the testing, the number of views is fixed at 12 and the number of points varies from 128 to 1,024. 128, 256, 384, 512, 640, 768, 896, and 1,024 points are respectively selected in the experiment. PVRNet is compared with state-of-the-art point cloud based model DGCNN. From the Fig. \ref{fig:rubust_of_point_num}, we can find that point missing greatly degrades the performance of DGCNN. When 256 views are used, the performance of DGCNN drops to about 50\%, while PVRNet can still keep a satisfactory performance of around 80\%. The relatively stable result benefits from the complement of corresponding view information which can help to compensate for the influence of missing points. 

The detailed experimental results demonstrate that our PVRNet is robust to the problem of data missing. That is due to the multimodal compensation of our framework. Point cloud feature and view feature are complementary to each other in the fusion process.

\section{Conclusions}
In this paper, we propose a point-view relation neural network called PVRNet for 3D shape recognition and retrieval. PVRNet is a novel multimodal fusion network which takes full advantage of the relationship between point cloud and views. In our framework, we fuse the view features and the point cloud feature in two branches. One branch uses the relation between the point cloud and each single view to generated a fusion feature called point-single-view fusion feature. The other branch exploits the relations between the point cloud and multiple sets of views, each of which has different number of most relevant views. Then, we further fuse the fusion features from these two branches to obtain a unified 3D shape representation. The effectiveness of our framework is evaluated by comprehensive experiments on ModelNet40 on the tasks of classification and retrieval. Ablation results and visualization results indicate that
the proposed point-view relation based feature fusion scheme has significant contribution to the proposed framework.

\section{Acknowledgments}
This work was supported by National Key R$\&$D Program of China (Grant No. 2017YFC0113000), and National Natural Science Funds of China (U1701262, U1705262, 61671267), National Science and Technology Major Project (No. 2016ZX01038101), MIIT IT funds (Research and application of TCN key technologies) of China, and The National Key Technology R$\&$D Program (No. 2015BAG14B01-02).
% AAAI is especially grateful to Peter Patel Schneider for his work in implementing the aaai.sty file, liberally using the ideas of other style hackers, including Barbara Beeton. We also acknowledge with thanks the work of George Ferguson for his guide to using the style and BibTeX files --- which has been incorporated into this document  --- and Hans Guesgen, who provided several timely modifications, as well as the many others who have, from time to time, sent in suggestions on improvements to the AAAI style. 

% The preparation of the \LaTeX{} and Bib\TeX{} files that implement these instructions was supported by Schlumberger Palo Alto Research, AT\&T Bell Laboratories, Morgan Kaufmann Publishers, The Live Oak Press, LLC, and AAAI Press. Bibliography style changes were added by Sunil Issar. \verb+\+pubnote was added by J. Scott Penberthy. George Ferguson added support for printing the AAAI copyright slug. Additional changes to aaai.sty and aaai.bst have been made by the AAAI staff.

\bibliography{mybib.bib}
\bibliographystyle{aaai}
\end{document}